\pdfoutput=1
\documentclass[11pt,a4paper]{article}
\usepackage[hyperref]{acl2018}
\usepackage{times}
\usepackage{latexsym}
\usepackage{url}
\usepackage{caption3} % load caption package kernel first
\DeclareCaptionOption{parskip}[]{} % disable "parskip" caption option
\usepackage[small]{caption}
\usepackage{amsmath, amssymb, amsthm}  % For mathematical symbols
\usepackage{graphics}
\usepackage[demo]{graphicx}

\aclfinalcopy % Uncomment this line for the final submission
\title{DeepEmo: Learning and Enriching Pattern-Based Emotion Representations}

\author{Elvis Saravia\\
  \normalsize National Tsing Hua University\\
  Hsinchu, Taiwan \\
  {\tt ellfae@gmail.com} \\\And
  Hsien-Chi Toby Liu \\
  \normalsize National Tsing Hua University\\
  Hsinchu, Taiwan \\
  {\tt tobbymailbox@gmail.com } \\\And
  Yi-Shin Chen \\
  \normalsize National Tsing Hua University\\
  Hsinchu, Taiwan \\
  {\tt yishin@gmail.com}
  }

\date{}

\begin{document}
\maketitle
\begin{abstract}
We propose a graph-based mechanism to extract rich-emotion bearing patterns, which fosters a deeper analysis of online emotional expressions, from a corpus. The patterns are then enriched with word embeddings and evaluated through several emotion recognition tasks. Moreover, we conduct analysis on the emotion-oriented patterns to demonstrate its applicability and to explore its properties. Our experimental results demonstrate that the proposed techniques outperform most state-of-the-art emotion recognition techniques.
\end{abstract}

\section{Introduction}
\label{sec:introduction}

Emotions can be defined as conscious affect attitudes, which constitute the display of a feeling. An emotion classification task consists of the representation learning or manual feature extraction of emotional words and phrases. Although there is constant debate about what exactly constitutes an emotion~\cite{weidman2017jingle}, there is no doubt of the societal and economic benefits that emotion recognition models and their applications can offer. Emotions are key influencers to understand other human social behaviors, such as \textit{motivation}, \textit{interest}, \textit{sarcasm}, and \textit{mental health}. Recently, emotion detection capabilities have been embedded into empathy-aware, AI conversational agents, such as Woebot~\footnote{\url{https://woebot.io/}} and in the dialogue system proposed by~\cite{zhou2017emotional}. The motivation of our work stems from the need to better model and explore different forms of online emotional expressions, particularly implicit ones. The proposed emotion representations allows emotion recognition systems to consider linguistic components such as stop words, which are usually ignored in emotion analysis but form an integral part of how we express our emotions and opinions ~\cite{pennebaker2007linguistic}.

Emotion recognition from text is challenging since emotional expressions can be highly implicit and are subject to evolve over time. This presents a challenge when relying on resources (e.g., emotion lexicons) that were generated by hand-crafted linguistic rules. For instance, mispronounced words appearing together will not be identified as the same when applying conventional feature extractors such as bag of words and n-grams. Another common tendency in online social networks is the use of different forms of expression, such as \textit{slang}, \textit{code words} and \textit{emoticons}, to express feelings and opinions. To address this problem, we design an algorithm, based on graph-theory, similar to~\cite{santos2017enriching}, to automate the process of extracting emotion representations. 

As an overview, we first collect an emotional corpus through noisy labels, which is then modeled via \textit{distant supervision} as in~\cite{go2009twitter}. Then, emotion features are extracted via a graph-based mechanism, which are further enriched with word embeddings in order to preserve semantic meaning between patterns. To evaluate the quality of patterns, emotion detection models are trained using various online classifiers and deep learning models. Our main contributions are summarized as follows: 1) A graph-based mechanism for automatic emotion-based feature extraction, 2) a set of emotion-rich feature representations used to conduct various emotion recognition tasks and other relevant target tasks, 3) a comprehensive performance analysis of various conventional learning models and deep learning models as it applies to emotion recognition from text, and 4) an emotion-rich lexicon, which is offered as open source, that allows for deeper analysis of a given emotion-relevant corpus.

\section{Related Work}
\label{sec:related-work} 

\subsection{Overview of Feature Representations}
\label{subsec:features}
We compare various feature extractors against the proposed technique, across two dimensions: 1) \textit{Coverage} - the features should be able to capture important implicit and explicit emotional information, and 2) \textit{Adaptability} - the features can apply to other type of emotional corpora, originating from different domains. Recent emotion recognition systems employ \textit{representation learning} for feature detection~\cite{poria2016convolutional, savigny2017emotion, nguyen2017sentence, abdul2017emonet}. In general, a combination of word embeddings (e.g., word2vec~\cite{mikolov2013efficient}) as input and a deep learning model, such as convolutional neural network (CNN), performs well for sentence classification~\cite{kim2014convolutional, zhang2015character, felbo2017using}. Due to the nature of these type of models and the type of features they learn, they tend to have high coverage, high adaptability, require little supervision (i.e., features are automatically learned), and capture context to some extent. However, there is a well-known trade-off between interpretability and high performance with these type of models. Our graph-based feature extraction mechanism focuses more on the underlying interaction between linguistic components. Therefore, the patterns automatically surface both implicit and explicit emotional expressions.

\subsection{Emotion Corpus and Models}
%One of the main challenges in text-based emotion-recognition studies is the acquisition and availability of emotional corpora. 

There are several open affective datasets, such as SemEval-2007 Affective Text Task~\cite{strapparava2007semeval} and Olympic games dataset~\cite{sintsova2013fine}. However, these emotion datasets are either limited by lack of fine-grained emotion labels or quantity. We bootstrap a set of noisy labels used to obtain larger collections of emotional tweets, and then perform annotations via distant supervision similar to~\cite{read2005using, go2009twitter, mintz2009distant, gonzalez2011identifying, mohammad2012emotional, purver2012experimenting, wang2012harnessing, mohammad2015using, abdul2017emonet}. In emotion recognition studies, the Plutchik's wheel of emotions~\cite{plutchik2001nature} or Ekman's six basic emotions~\cite{ekman1992argument}, are commonly adopted to define emotion categories~\cite{mohammad2012emotional, suttles2013distant}. Emoticons and emojis have also proven to be useful for defining emotion categories~\cite{eisner2016emoji2vec, felbo2017using}. Similar to~\cite{mohammad2015using, liew2016exploring, abdul2017emonet}, we rely on hashtags to define our emotion categories. 

\subsection{Emotion Lexica}
Emotion classifiers have enabled understanding of mood patterns displayed by mental health patients~\cite{park2012depressive, de2013predicting, harman2014measuring, coppersmith2014quantifying}. Some of these studies rely on a predefined lexicon, such as \textit{LIWC}~\cite{pennebaker2007linguistic}~\footnote{LIWC stands for linguistic inquiry and word count}, \textit{WordNet Affect}~\cite{strapparava2004wordnet} and \textit{EmoLex}~\cite{mohammad2013crowdsourcing}, to extract emotional cues from text-based corpora. A recent study demonstrates the correlation between emotional tone and perceived demographic traits among users in a social network~\cite{volkova2016inferring}. This study relies on an emotion detection system, which is built using lexical features, such as \textit{emoticons} and \textit{hashtags}~\cite{pang2002thumbs}. Other user information, such as age and gender were obtained from external sources, which limited the amount of data that the authors could collect. An improvement to their work would be to use the content from the users' tweets to automatically determine user attributes, such as age and gender~\cite{sap2014developing}. Other works use hand-crafted linguistic features to improve emotion classification performance~\cite{blitzer2007biographies, wang2012harnessing, roberts2012empatweet, qadir2013bootstrapped, volkova2013exploring, mohammad2015using, volkova2016inferring, becker2017multilingual}. These features are useful for emotion classification but offer limited coverage. Our emotion lexicon is constructed with an emphasis on \textit{coverage} (i.e., captures implicit and explicit emotional expressions).

\section{Methodology}
\label{sec:approach}

\subsection{Graph-Based Representations}
\label{subsec:graph-rep}
In this section, we introduce a graph-based feature extraction algorithm, which automatically extracts a set of emotion-rich syntactic patterns. For notation purposes, we denote scalars with italics (e.g., \textit{u}), vectors with bold lowercase (e.g., $\mathbf{v}$), and matrices with bold uppercase (e.g., $\mathbf{X}$). The patterns $P = \{ p_1, p_2, ...,p_n \}$ will be assigned a weight, also referred to as a \textit{pattern score}, which is used to determine how important a pattern $p$ is to an emotion $e$. In the context of an emotion classifier, patterns and their weights play the role of features. The graph-based feature extraction algorithm is summarized in the following steps:

\textbf{Step 1 (Normalization):} First, two separate collection of documents -- subjective tweets $S$ (obtained through hashtags as noisy labels) and objective tweets $O$ (obtained from news accounts) -- are obtained using the Twitter API~\footnote{Each dataset contains over 2+ million tweets.}. Both datasets are tokenized by white-spaces and then further preprocessed by applying lower case and replacing user mentions and URLs with a $<$\textit{usermention}$>$ and $<$\textit{url}$>$ placeholder, respectively. Hashtags, are used to obtain ground-truth in this work, so to avoid any bias we replace them with $<$\textit{hashtag}$>$.

\textbf{Step 2 (Graph Construction):} Given the normalized objective tweets $O$ and subjective tweets $S$, two graphs are constructed: objective graph $G_{o}(V_o; A_o)$ and subjective graph $G_{s}(V_s; A_s)$, respectively. Vertices $V$ is a set of nodes which represent the tokens extracted from the corpus. Edges, denoted as $A$, represent the relationship of words as extracted from a piece of text using a window approach. This consideration is important as it preserves the prosody and underlying syntactic structure of textual data. For instance, a post ``$<$usermention$>$ last night's concert was just awesome !!!!! $<$hashtag$>$" results in the following set of arcs: \textit{``$<$usermention$>$ $\rightarrow$ last'', ``last $\rightarrow$ night'', ... , ``!!!!! $\rightarrow$ $<$hashtag$>$''}.

\textbf{Step 3 (Graph Aggregation):} The goal of this step is to obtain a set of arcs that are more relevant to subjectivity or emotional expressions. The assumption is that by adjusting graph $G_s$ with $G_o$ it is possible to obtain a new graph $G_e$, also referred to as an emotion graph. $G_e$ preserves emotion-relevant tokens, which is achieved in two steps: 

(1). For an arc $a_i \in A$, its normalized weight can be computed as shown in Equation~\ref{weight}.

\begin{equation}
\label{weight}
 w(a_i) = \frac{freq(a_i)}{\max_{j \in A} freq(a_j)}
\end{equation}

where $freq(a_i)$ is the frequency of arc $a_i$.

(2). Subsequently, new weights for arcs $a_i \in G_e$ are assigned based on a pairwise adjustment as shown in Equation~\ref{uweight}.
    
\begin{equation}
\label{uweight}
w(a_{i})= 
\begin{cases}
  w(a_{s_i}) - w(a_{o_j}), & \text{if }
       a_{o_j} = a_{s_i} \in G_o
       
\\
 w(a_{s_i}), & \text{otherwise}
\end{cases}
\end{equation}

The resulting weights belonging to graph $G_e$ were adjusted so that the most frequently occurring arcs in objective set $G_o$ are weakened in $G_e$. As a result, arcs in $G_e$ that have higher weights represent tokens that are more relevant to subjective content. Furthermore, arcs $a_i \in A_e$ are pruned based on a threshold $\phi_w$~\footnote{$\phi_w$ is an experimentally defined threshold.}.

\textbf{Step 4 (Token Categorization):} Given an adjacency matrix $\mathbf{M}$, an entry $M_{i,j}$ is computed as:

\begin{equation}
\label{adjacency}
    M_{i,j}= 
    \begin{cases}
        1 & \mbox{ if node $i$ and $j$ are linked in } G_e\\
        0 & \mbox{ otherwise}
    \end{cases}
\end{equation}

Then, eigenvector centrality and clustering coefficient of all vertices in $V_e$ are computed, which will be used to categorize tokens into two types: \textit{connector words} and \textit{subject words}. 

(1) \textbf{Connector Words:} To measure the influence of all nodes in graph $G_e$, we utilize eigenvector centrality, which is computed as:

\begin{equation}
\label{centrality_eq}
c_i = \frac{1}{\lambda}\sum_{\substack{j\in V_e \\}}M_{i,j}c_j
\end{equation}

where $\lambda$ denotes a proportionality factor and $c_i$ is the centrality score of node $i$. 

Given $\lambda$ as the corresponding eigenvalue, Equation~\ref{centrality_eq} can be reformulated in vector notation form as $\mathbf{M}\mathbf{c} = \lambda \mathbf{c}$, where $\mathbf{c}$ is an eigenvector of $\mathbf{M}$. Given a selected eigenvector $\mathbf{c}$ and the eigenvector centrality score of node $i$, denoted as $c_i$, the final list of connected words, hereinafter referred to as $CW$, is obtained by retaining all tokens with $c_i > \phi_{eig}$~\footnote{$\phi_{eig}$ is an experimentally defined threshold.}. $CW$ represents the set of words that are very frequent and contain high centrality (e.g., ``or", ``and", and ``my"). 

(2) \textbf{Subject Words} In contrast, subject words or topical words are usually clustered together, i.e., many subject words are interconnected by the same connector words. Therefore, a coefficient is assigned to all nodes in $G_e$ and is computed as:

\begin{equation}\label{ind_clust_eq}
cl_i = \frac{\sum_{\substack{j \neq i; k \neq j; k \neq i}}M_{i,j} \times M_{i,k} \times M_{j,k}}{\sum_{\substack{j \neq i; k \neq j; k \neq i}}M_{i,j} \times M_{i,k}} \times \frac{1}{|V_e|}
\end{equation}

where $cl_i$ denotes the average clustering coefficient of node $i$, which captures the amount of inter-connectivity among neighbours of node $i$. 
Similar to the connector words, the subject words, hereinafter referred to as $SW$, are obtained by retaining all the tokens with $cl_i > \phi_{cl}$~\footnote{$\phi_{cl}$ is an experimentally define threshold.} Examples of subjects words are (e.g., ``never" and ``life" ).

\textbf{Step 5 (Pattern Candidates):} Given the set of tokens, $SW$ and $CW$, we employ a bootstrap approach to construct candidate patterns which express subjective meaning without losing syntactic structure. Consequently, the following are some of the rules which are used to define the \textit{candidate patterns}: $<sw, sw, cw>$, $<sw, cw, sw>$, $<cw, sw, sw>$, and $<cw, cw, sw>$, where $sw$ and $cw$ represent arbitrary tokens obtained from the set $SW$ and $CW$, respectively. It's important to clarify that sequences of size two and three were used in this work since this setting experimentally worked best for us. We may sometimes refer to these candidate patterns as \textit{templates}, similar to~\cite{riloff1996automatically, riloff2003learning, tromp2015pattern}. The difference in our work is that we don't impose grammatical heuristics or rules in the pattern extraction process, therefore, our patterns tend to naturally have higher coverage and capture both implicit and explicit emotional content.

\textbf{Step 6 (Basic Pattern Extraction):} A naive pattern extraction process consists of applying the syntactic templates to a training corpus~\footnote{Subjective dataset $S$ is used to process the templates.} in an exhaustive manner. In addition, subject words $sw$ in each pattern is replaced with a $<$\textbf{*}$>$ placeholder. This operation allows for unknown subject words, not present in our training corpus, to be considered when modeling on an external emotional corpus. We are interested in patterns that are highly associated with subjectivity, so patterns frequently occurring above a threshold are kept and the rest are filtered out.~\footnote{A grand total of 19,821 patterns were extracted.} In Table~\ref{table:patterns_sample}, we provide examples of the type of basic patterns extracted along with the corresponding templates. Next, we discuss the process of enriching the syntactic patterns with word embeddings. This enrichment process helps to preserve semantic between patterns and improves feature relevance~\cite{santos2017enriching}.

%(The complete set of basic patterns per emotion category are provided as supplementary material)

\begin{table}
\centering
\small
\begin{tabular}{|c|c|}
  \hline
  \textbf{Templates} &\textbf{Pattern Examples} \\
  \hline
  $<cw, sw>$& ``stupid *" , ``like *", ``am *"  \\
  $<cw, cw, sw>$&``love you *", ``shut up *"  \\
  $<sw, cw, sw>$&``* for *"\\
  $<sw, cw, cw>$ & ``* on the" , ,   \\
  $<sw, cw>$&``* $<$hashtag$>$"  \\
  \hline
\end{tabular}
\caption{Examples of patterns and templates extracted through the basic pattern extraction mechanism.}
\label{table:patterns_sample}
\end{table}

\subsection{Enriched Patterns}
\textbf{Weighted Word Embeddings} First, we obtain pre-trained Twitter-based word embeddings from~\cite{deriu2017leveraging} and reweigh them via a sentiment corpus through distant supervision~\cite{read2005using, go2009twitter}~\footnote{We collected approximately 10 million tweets via sentiment emoticons (5+ mil negative and 5+ mil positive).}. We trained a fully connected deep neural network with 10 epochs (1 hidden layer) via backpropagation as in~\cite{deriu2017leveraging}. We denote the sentiment word embeddings as $\mathbf{W}\in \mathbb{R}^{d\times n}$ where $d=52$. Note that term frequency-inverse document frequency (\textit{tf-idf}) was used to reduce the vocabulary of words (from 140K to 20K words).

\textbf{Word Clusters} We then apply agglomerative clustering to generate clusters of semantically related words through their word embedding information. To determine the quality of the clusters they are compared with \textit{WordNet-Affect} synsets~\cite{strapparava2004wordnet} and tested for both \textit{homogeneity} and \textit{completeness}. We use \textit{Ward's method}~\cite{ward1963hierarchical} as the linkage criterion and \textit{cosine distance} as the distance metric. In the end, we obtained $k=1500$ clusters. We use the scikit-learn implementation to perform the word clustering (\url{http://scikit-learn.org}).

\textbf{Enriched-Pattern Construction} The purpose of the word clusters is to use them to guide the process of enriching the patterns. In other words, the patterns will hold some semantic relationship, which becomes useful for classification problems. Note that this process is similar to the naive pattern extraction with the exception of the word embedding integration. This entails a bootstrap process where an emotional corpus is processed and candidate patterns are searched in an exhaustive fashion. Any word sequences in the emotional corpus that satisfies the templates are retained and the rest are filtered out. In addition, the $sw$ component of the templates must be a word found in the word clusters defined above. Furthermore, patterns that appear $<10$ are filtered out, producing a total of 187,647 patterns. In Section~\ref{sec:analysis}, we analyze the patterns more in depth and provide examples.

\subsection{Emotion Pattern Weighing}
\label{subsec:weighing}
The patterns extracted in the previous step are still not mapped to any specific emotion category. Before training a classification model, a \textit{pattern weighing mechanism} needs to be employed. Similar to other popular weighing mechanisms, such as \textit{tf-idf}, weights determine the importance of patterns to each emotion $e_j \in E$. The proposed pattern weighing scheme is a modification of \textit{tf-idf}, coined as \textit{pattern frequency-inverse emotion frequency} (\textit{pf-ief}), and is defined in two steps. Firstly, we compute for $pf$ as:

\begin{equation}
\label{pattern_frequency}
%pf_{p,e} = log (freq(p,e) + 1)
pf_{p,e} = \log {\frac {\sum\limits_{p_i\in P_e}{freq(p_i,e)} + 1}{freq(p,e)+1}}
\end{equation}

where $freq(p,e)$ represents the frequency of $p$ in $e$, and $pf_{p,e}$ denotes the logarithmically scaled frequency of a pattern $p$ in a collection of texts related to emotion $e$, 

Then we compute for $ief$ as:
\begin{equation}
\label{inverse_emotion_frequency}
%ief_p = \frac{|E|}{|\{e \in E : freq(p,e) > 0\}|}
ief_p = \log{\frac{freq(p,e)+1}{\sum\limits_{e_j \in E}{freq(p,e_j)}+1}}
\end{equation}

where the inverse emotion frequency $ief_p$ is a measure of the relevance of pattern $p$ across all emotion categories.

Finally, we obtain a pattern score as:

\begin{equation}
\label{emotion_degree}
 ps_{p,e} = pf_{p,e} \times ief_p % \times cov_p
\end{equation}
where $ps_{p,e}$ is the final score that reflects how important a pattern $p$ is to an emotion class $e$.

\section{Models}
\label{sec:methodology}
\subsection{DeepEmo}
The proposed framework, coined as \textbf{DeepEmo}, combines a multilayer-layer CNN architecture with a matrix form of the proposed graph-based features. The input $\mathbf{X}\in \mathbb{R}^{n\times m}$ denotes an embedding matrix where entry $X_{i,j}$ represents the pattern score of enriched pattern $i$ in emotion $j$.~\footnote{We use a zero-padding strategy to adjust the embeddings as in~\cite{kim2014convolutional}} The input is fed into 2 1d convolutional layers with filters of size $3$ and $16$. The output of this process is passed through a non-linear activation function (i.e., ReLU~\cite{nair2010rectified}) and produces a feature map matrix. A 1-max pooling layer~\cite{boureau2010theoretical} of size 3 is then applied to each feature map. The results of the pooling are fed into two hidden layers of dimensions $512$ and $128$ in that order, each applied a dropout~\cite{hinton2012improving} of $0.8$ for regularization. We chose a batch size of $128$ and trained for $7$ epochs using Adam~\cite{kingma2014adam} optimizer. A softmax function is used to generate the final classification. We use Keras~\cite{chollet2015keras} to implement the CNN architecture.

\subsection{Vector Model}
As a baseline, we present a naive vector model (\textbf{EVM}), which demonstrates basic usability and applicability of the basic patterns proposed in Section~\ref{subsec:graph-rep}. Pattern weights are obtained using the pattern weighing mechanism proposed in Section~\ref{subsec:weighing}. Formally, given $n$ patterns and $m$ emotions, we can represent the entire emotion model as matrix $\mathbf{EM}\in\mathbb{R}^{n\times m}$. An entry $EM_{i,j}$ represents the rank of basic pattern $i$ in emotion $j$, which is based on the pattern score $ps_{i,j}$. Note that patterns with higher $ps$ values have lower rank values, as in they are more relevant to that particular emotion. Assume a social post $tw$ for which we want to obtain its portrayed emotion, we first compute its frequency vector $\mathbf{f}\in \mathbb{R}^n$, where entry $f_i$ represents the frequency of pattern $i$ in input social post $d$. We compute the emotion scores as:

\begin{equation}
 \mathbf{es} = \mathbf{f} \cdot   \mathbf{EM}  %= 
\end{equation}

where $\mathbf{es}\in \mathbb{R}^m$ and entry $es_{j}$ corresponds to the final emotion score of emotion $j$ for the post $tw$. The index of the minimum of these values is selected as the final emotion detected for $tw$.

\begin{table*}[!ht]
\small
\centering
\tabcolsep=0.11cm
    \begin{tabular}{lclclcccccccccl}
        \textbf{Models}&\textbf{Features} & \textbf{anger} & \textbf{anticipation} & \textbf{disgust} & \textbf{fear} & \textbf{joy} & \textbf{sadness}  & \textbf{surprise} & \textbf{trust} &\textbf{F1 Avg.}
        \\ \hline
                    
        \multicolumn{1}{l|}{\textbf{BoW}} & \multicolumn{1}{l|}{\textbf{word frequency}}  & 0.53 & 0.08  & 0.17  & 0.53  & 0.71 & 0.60 & 0.36  & 0.33 & 0.57
        \\ 
        \multicolumn{1}{l|}{\textbf{BoW\textsubscript{TF-IDF}}}&
        \multicolumn{1}{l|}{\textbf{TF-IDF}}  & 0.55 & 0.09  & 0.18  & 0.57 & 0.73 & 0.62  & 0.39 & 0.35 & 0.60 
        
        \\\hline
        \multicolumn{1}{l|}{\textbf{n-gram}}&
        \multicolumn{1}{l|}{\textbf{word frequency}}  & 0.56 & 0.09 & 0.17  & 0.57  & 0.73 & 0.64 & 0.42  & 0.39 & 0.61
        
        \\
        \multicolumn{1}{l|}{\textbf{n-gram\textsubscript{TF-IDF}}}&
        \multicolumn{1}{l|}{\textbf{TF-IDF}}  & 0.58 & 0.12  & 0.17  & 0.60  & \textbf{0.75} & 0.67  & 0.47 & 0.45 & 0.63
        
        \\\hline
        \multicolumn{1}{l|}{\textbf{char}}&
        \multicolumn{1}{l|}{\textbf{character frequency}} & 0.35 & 0.03 & 0.04 & 0.20& 0.51& 0.46 & 0.10  & 0.12 & 0.37
        \\
        \multicolumn{1}{l|}{\textbf{char\textsubscript{TF-IDF}}}&
        \multicolumn{1}{l|}{\textbf{TF-IDF}}  & 0.33 & 0.03 & 0.06  & 0.21 & 0.52 & 0.45  & 0.11 & 0.13 & 0.37 
        
        \\\hline
        \multicolumn{1}{l|}{\textbf{char\_ngram}}&
        \multicolumn{1}{l|}{\textbf{character frequency}}  & 0.49 & 0.06 & 0.12  & 0.46  & 0.67 & 0.55 & 0.30  & 0.28 & 0.52
        
        \\
        \multicolumn{1}{l|}{\textbf{char\_ngram\textsubscript{TF-IDF}}}&
        \multicolumn{1}{l|}{\textbf{TF-IDF}}  & 0.53 & 0.07 & 0.15  & 0.53 & 0.71 & 0.59  & 0.35 & 0.31 & 0.57

        \\\hline
        \multicolumn{1}{l|}{\textbf{word2vec}}&
        \multicolumn{1}{l|}{\textbf{word embeddings}} & 0.50 & 0.02 & 0.13 & 0.48& 0.69& 0.51& 0.35 &0.31&0.53

        \\\hline
        \multicolumn{1}{l|}{\textbf{LIWC}}&
        \multicolumn{1}{l|}{\textbf{affect words}} & 0.35 & 0.03 & 0.11& 0.30& 0.49& 0.35& 0.18 & 0.19&0.35
        
        \\\hline
        \multicolumn{1}{l|}{\textbf{EVM}}&
        \multicolumn{1}{l|}{\textbf{patterns}} & 0.42 & 0.02 & 0.04& 0.38& 0.50& 0.34& 0.24 &0.21&0.38
        \\\hline
        \multicolumn{1}{l|}{\textbf{CNN-patt}}&
        \multicolumn{1}{l|}{\textbf{basic patterns}} & 0.47 & 0.00 & 0.00& 0.45& 0.67& 0.61& 0.15 & 0.08&0.52
        \\
        %\hline
        \multicolumn{1}{l|}{\textbf{DeepEmo}}&
        \multicolumn{1}{l|}{\textbf{enriched patterns}} & \textbf{0.58} & \textbf{0.16} & \textbf{0.32}& \textbf{0.65}& \textbf{0.75}& \textbf{0.70}& \textbf{0.59} & \textbf{0.55}&\textbf{0.67}

    \end{tabular}
    \caption{Comparison of our model against conventional feature extractors using F1-score. \textbf{LIWC} uses a bag of words approach. \textbf{word2vec} model adopts pre-trained embeddings from~\cite{mikolov2013efficient}. \textbf{char} refers to character-level features. \textbf{n-gram} employ unigrams, bigrams, and trigrams as features. \textbf{CNN-patt} uses the proposed CNN architecture with basic patterns.}
    \label{f1-table}
\end{table*}

\subsection{Comparison Models}
\subsubsection{Traditional models}
We compare DeepEmo against various traditional methods (e.g., bag of words (\textbf{BoW}), character-level (\textbf{char}), \textbf{n-grams}, \textbf{TF-IDF}) commonly used in sentence classification. The classifier used to train these models is the stochastic gradient descent (SGD) classifier provided by scikit-learn.

\subsubsection{Deep Learning models}
Deep learning architectures enable automatic learning of features from textual information. We observed that among the works that employ deep learning models for emotion classification, they vary by the choice of input: \textit{pre-trained word/character embeddings} and \textit{end-to-end learned word/character representations}. Our work differs in that we utilize enriched graph-based representations as input, therefore, we believe it is also important to compare with these methods. We compare with convolution neural networks (\textbf{CNNs}), recurrent neural networks (\textbf{RNNs}), bidirectional gated recurrent neural networks (\textbf{GRNNs}), and word embeddings (\textbf{word2vec})~\cite{mikolov2013efficient}.

%~\footnote{All parameters are provided in supplementary materials.}

\begin{table}[!ht]
\small
\centering
    
    \tabcolsep=0.11cm
    \scalebox{0.85}{
    \begin{tabular}{|c|c|c|c|}
    \hline
    \textbf{Emotions} & \textbf{Train} & \textbf{Test} & \textbf{Hashtags}\\
    \hline
    \textbf{sadness} & 192842  & 21422  & \#depressed, \#grief  \\ 
    \hline
    \textbf{joy} & 149986  & 16663  & \#fun, \#joy \\
    \hline
    \textbf{fear}  & 92145  &  10209  & \#fear, \#worried\\
    \hline
    \textbf{anger}  &  91947 & 10200 & \#mad, \#pissed\\
    \hline
    \textbf{surprise}  & 41337  & 4691  & \#strange, \#surprise\\
    \hline
    \textbf{trust}  & 17295  & 1913 & \#hope, \#secure\\
    \hline
    \textbf{disgust}  & 8052  & 873 & \#awful, \#eww\\
    \hline
    \textbf{anticipation}  & 3588 & 384 & \#pumped, \#ready\\
    \hline
    \end{tabular}}
    \caption{Distribution of train and test datasets.}
    \label{emotion_distribution_training}
\end{table}

\section{Experiments}

\begin{table*}[!ht]
\small
\centering
    
    \tabcolsep=0.11cm
    \scalebox{0.85}{
    \begin{tabular}{|c|c|c|c|c|}
    \hline
    \textbf{Model} & \textbf{Adopted from} &\textbf{Input} &\textbf{Epochs} & \textbf{Accuracy}\\
    \hline
    \textbf{RNN} & ~\cite{husein} & word2vec~\cite{mikolov2013efficient} & 24  & 0.53 \\
    \hline
    \textbf{CNN} & ~\cite{kim2014convolutional} & character embeddings (end-to-end) & 50 & 0.63 \\
    \hline
    \textbf{Bi-GRNN} & ~\cite{ilya2017} &enriched patterns (ours) & 12  & \textbf{0.65} \\
    \hline
    \end{tabular}}
    \caption{Results of \textbf{DeepEmo} against other deep learning models adopted and modified to perform emotion classification.}
    \label{deep-learning}
\end{table*}

\subsection{Data}
\label{subsec:emotional-corpus}
%One of the requirements for training deep neural networks is the availability of large-scale datasets. 
We follow~\cite{mohammad2012emotional, wang2012harnessing, abdul2017emonet} and construct a set of hashtags (grounded on Plutchik's wheel of emotions~\cite{plutchik2001nature}) to collect English tweets from Twitter API. Specifically, we use the eight basic emotions of Plutchik: \textit{anger}, \textit{anticipation}, \textit{disgust}, \textit{fear}, \textit{joy}, \textit{sadness}, \textit{surprise}, and \textit{trust}. The hashtags serve as noisy labels, which allows annotation of the data through distant supervision~\cite{go2009twitter}. 339 hashtags were defined in total. To ensure tweets quality, we follow pre-processing steps proposed by~\cite{abdul2017emonet} and considered the hashtag appearing in the last position of a tweet as the ground truth. We split the data into training (90\%) and testing (10\%). The final distribution of the data and a list of hashtag examples for each emotion are provided in Table~\ref{emotion_distribution_training}. In the following sections, we evaluate the effectiveness of the enriched patterns on several emotion recognition tasks. We use F1-score as the evaluation metric, which is commonly used in emotion recognition studies due to the imbalanced nature of the emotion datasets.

\subsection{Experimental Results}
\textbf{Traditional Feature Extractors} The results obtained from the traditional feature extractors are presented in Table~\ref{f1-table}. As the table shows, TF-IDF models usually produce better results than basic count-based features for both character-level and word-level feature extractors. These findings are consistent with the work of~\cite{zhang2015character}, where traditional methods, such as n-gram TF-IDF, were found to perform comparable to neural networks on various sentence classification tasks. 

\textbf{Results with Pattern Approaches} The results of \textbf{EVM} and \textbf{CNN-patt}, which employ the \textit{basic graph-based patterns}, are worst that most of the conventional approaches. \textbf{DeepEmo}, which uses the \textit{enriched patterns}, acquires better results (F1-score of 67\%) than both \textbf{CNN-patt} and \textbf{EVM}, and all of the other conventional approaches. In fact, our method obtains the best F1-score on all emotions. We can also observe that there is a significant boost in performance (+15\%) when using the enriched patterns (DeepEmo) as compared to the basic patterns model (CNN-patt). Overall, we can observe that the enriched graph-based features are feasible for training emotion recognition models. 

\textbf{Comparison to state-of-the-art} We also compare results with published literature, which employ emotion recognition systems using Ekman's six basic emotions. For fair comparison, we reduced our dataset from eight emotions to six emotions: \textit{anger}, \textit{disgust}, \textit{fear}, \textit{joy}, \textit{sadness}, and \textit{surprise}. As shown in Table~\ref{published-comparison}, our emotion recognition system achieves better results (F1-score of 0.72\%) than most of the methods with the exception of~\cite{volkova2016inferring}. Their emotion recognition system performs better than ours (F1-score of 78\%) since they use well-defined linguistic features, such as emoticons and hashtags. Our features are more susceptible to noise because we aim for higher coverage in order to capture more implicit emotional expressions. This consideration is important if we intend to use the emotion lexicons for conducting deep analysis on affective datasets. In addition, their features are domain-specific, which means some important features, such as emoticons and hashtags, may not be applicable to other affective datasets. According to~\cite{zhang2015character}, traditional methods are strong candidates on this type of tasks for dataset of size up to the hundreds of thousands, and only after several millions do CNN models start to do better. We plan to continue enlarging our datasets and refining pattern weights, which are feasible methods to improve results.

\begin{table}[!ht]
\small
\centering
    
    \tabcolsep=0.11cm
    \scalebox{0.85}{
    \begin{tabular}{|c|c|c|}
    \hline
    \textbf{Method} & \textbf{Data Size} & \textbf{F1-score}\\
    \hline
    \textbf{Roberts (2012)} & 3777   & 0.67 \\
    \hline
    \textbf{Qadir (2013)} & 4500 & 0.53 \\ 
    \hline
    \textbf{Mohammad (2015)}  & 21,051 &  0.49 \\
    \hline
    \textbf{Volvoka (2016)}  & 52,925  & \textbf{0.78} \\
    \hline
    \textbf{DeepEmo (Ours)}  & 597,192 & \textbf{0.72} \\
    \hline
    \end{tabular}}
    \caption{F1 average comparison of our method against other notable published literature.}
    \label{published-comparison}
\end{table}

\textbf{Results with Deep Learning} We offer a comparison with various deep learning models as evaluated on Ekman's six basic emotions. The architectures were adopted from published resources. We feed the enriched patterns as embeddings to a Bidirectional GRNN (adopted from~\cite{ilya2017}), and achieve the best results (Accuracy of 0.65\%) among the deep learning models, as shown in Table~\ref{deep-learning}. The results show that the enriched patterns can also be applied to other deep learning models besides CNNs, which leaves an opportunity for further exploration and experimentation.

\textbf{Affective Dataset} We conducted experiments on other existing affective datasets using the enriched patterns. We acquire better results (F1-score of 0.48\%) on SemEval-2007 Affective Text Task~\cite{strapparava2007semeval} as compared to the work of~\cite{felbo2017using}, which to the best of our knowledge, holds state-of-the-art results (37\%) on this dataset. We directly used their benchmark dataset and modified our models to support the available emotion labels. On the SemEval-2017 Task 4 we acquire an F1-score of 53\%. These results provide more evidence that our enriched patterns are applicable and \textit{adaptable} to other emotion-relevant tasks and datasets. 

\section{Analysis of Enriched Patterns}
\label{sec:analysis}

\begin{table*}[!ht]
\small
\centering
    \tabcolsep=0.11cm
    \scalebox{0.85}{
    \begin{tabular}{|c|c|c|c|c|c|}
    \hline
    \textbf{Emotion Dataset} & \textbf{Study}& \textbf{Task} & \textbf{Domain} & \textbf{Dataset Size} & \textbf{Enriched Patterns}\\
    \hline
    \textbf{Our Full Dataset} & Ours & Emotion (8) & Tweets & 1,896,849  &0.94 \\
    \hline
    \textbf{Gender Data} & Ours & Emotion (8)& Tweets &294,792  & 0.89 \\ 
    \hline
    \textbf{SemEval07 Task 14} & \cite{strapparava2007semeval} & Emotion (3) & Headlines & 601  &0.62 \\
    \hline
    \textbf{SemEval17 Task 4}  & \cite{SemEval:2017:task4} & Sentiment (3) & Tweets &20,621  & 0.99 \\
    \hline
    \textbf{SemEval18 Task 1}  & \cite{mohammad2017emotion}  & Emotions (4)* & Tweets &3890& 0.92 \\
    \hline
    \textbf{SST-2} & \cite{socher2013recursive}& Sentiment (5) & Reviews &58,990  &0.76 \\
    \hline
    \textbf{SST-5} & \cite{socher2013recursive}& Sentiment (5) & Reviews & 96,660  &0.71 \\
    \hline
    \textbf{PsychExp} & \cite{wallbott1988universal}& Emotion (5) & Experiences &7339  &0.95 \\
    \hline
    \end{tabular}}
    \caption{Statistics on word coverage per text of the enriched patterns on several affective datasets. * denotes that we used the \textbf{El-oc} testing data. The numbers inside the () represent the number of classes present in the dataset.}
    \label{coverage}
\end{table*}

\begin{table*}[!ht]
\small
\centering
    
    \tabcolsep=0.11cm
    \scalebox{0.85}{
    \begin{tabular}{|c|c|c|}
    \hline
    \textbf{Emotions} & \textbf{Male patterns}  & \textbf{Female patterns}\\
    \hline
    \textbf{Anger} & a\{crazy\}, you\{despise\}, like\{try\}  & my\{yelling\}, would \{want\}, hate \{you\} \\
    \hline
    \textbf{Sadness} & your \{lyrics\}, \{bouncing\} your & better \{come\}, you \{wreck\}, \{despise\} going \\ 
    \hline
    \textbf{Surprise} & last \{second\}, to \{announce\}  &  happy \{birthday\}, \{only\} person \\
    \hline
    \textbf{Fear}  & \{you\} have,  \{getting\} dark &   my \{stepmom\}, the \{loneliest\} \\
    \hline
    \end{tabular}}
    \caption{Examples of the top 1000 most frequently occurring patterns by gender.}
    \label{gender}
\end{table*}

In this section, we explore the enriched patterns extracted from a gender-based dataset. We collected user feeds from Twitter and classified users into male and female classes based on their content via Sap et al.'s gender predictor~\cite{sap2014developing}. This produces a gender dataset, which we also manually verify by ourselves. We randomly sampled 2000 males and 2000 females and then randomly sampled 100 tweets from each user feed. This generated 400,000 tweets in total, which we further reduced by filtering out tweets with $\leq5$ words. The final amount of tweets is 294,792, which we classify using DeepEmo. 

We apply a pattern frequency analysis on the gender data using the enriched patterns. The patterns that are shared by both males and females are discarded and the 1000 most frequently occurring patterns for each gender dataset are analyzed. Examples of the most frequent emotional patterns captured by $<cw, sw>$ and $<sw, cw>$ templates as expressed by both females and males are provided in Table~\ref{gender}. The words inside the $\{\}$ represent the subject words captured by the pattern enrichment process. We can observe that subject words represent emotion-rich words such as ``despise", ``yelling", and ``loneliest". The connecting words, on the other hard, provide context, which helps to better understand the enriched patterns.

We are currently investigating whether there are gender-specific emotional patterns or expressions on social media. However, it is too early to derive conclusions from the primitive analysis presented here. We can still observe that providing context helps to tell a story behind the emotional expressions. Another interesting research direction would be to use the patterns directly for gender prediction. The goal of the analysis was to explore the enriched patterns and show how they may be used for conducting deeper analysis on an emotional corpus.

\textbf{Pattern Coverage} We computed the coverage of the enriched patterns on several affective datasets. As shown in Table~\ref{coverage}, 89.4\% of the tweets in the gender data contains at least one of the enriched patterns. Our patterns also show high coverage on datasets from different domains, such as \textbf{SST-2} (76\%), \textbf{SST-5} (71\%)~\cite{socher2013recursive}, and \textbf{PsychExp} (95\%)~\cite{wallbott1988universal}. We observed that the dataset size did not influence the coverage results. A high coverage (95\%) was obtained on emotional experiences described in~\cite{wallbott1988universal}, which originate from a different domain from which the patterns were constructed. This shows that our enriched patterns are adaptable to other domains, which open opportunities for further exploration and experimentation.

%\begin{figure}
%    \centering
%    
%    \includegraphics[width=1\linewidth]{anger.png}
%    
%    \caption{Emotional expressions by gender. $\{G\}$ denotes subject words filtered in the word clustering step.}
%    \label{fig:emocloud}
%\end{figure}

\section{Discussion}
\label{sec:discussion}
\citet{abdul2017emonet} showed that improving data quality is an important step in improving emotion classification results (achieves an F1-score of 83\%). We observed that they report a larger dataset (790,059) and more balanced data collections for each emotion. In contrast, our dataset is more imbalanced, but even when balancing the results did not improve significantly (average F1-score of 68\%). At the time of writing this manuscript, the authors were still working on making their datasets publicly available, so unfortunately we couldn't compare directly with their method. As future work, we hope to keep refining our hashtags and improving the emotional corpus. All benchmark datasets, lexicons, pre-trained models, and code for running the models will be made available soon.

\section{Conclusion}
We proposed an enriched graph-based feature extraction mechanism to extract emotion-rich representations. The patterns are enriched with word embeddings and are used to train several effective emotion recognition models. Our patterns capture implicit emotional expressions which improves emotion recognition results and helps with interpretability. We demonstrate a basic application of the proposed affective lexicon on a gender dataset. We hope to improve the pattern weighing mechanism so as to improve the performance on emotion recognition tasks and minimize trade-off between pattern coverage and performance.

%\bibliography{acl2018}
\bibliography{acl2018}

\begin{thebibliography}{57}
\expandafter\ifx\csname natexlab\endcsname\relax\def\natexlab#1{#1}\fi

\bibitem[{Abdul-Mageed and Ungar(2017)}]{abdul2017emonet}
Muhammad Abdul-Mageed and Lyle Ungar. 2017.
\newblock Emonet: Fine-grained emotion detection with gated recurrent neural
  networks.
\newblock In \emph{Proceedings of the 55th Annual Meeting of the Association
  for Computational Linguistics (Volume 1: Long Papers)}, volume~1, pages
  718--728.

\bibitem[{Becker et~al.(2017)Becker, Moreira, and dos
  Santos}]{becker2017multilingual}
Karin Becker, Viviane~P Moreira, and Aline~GL dos Santos. 2017.
\newblock Multilingual emotion classification using supervised learning:
  Comparative experiments.
\newblock \emph{Information Processing \& Management}, 53(3):684--704.

\bibitem[{Blitzer et~al.(2007)Blitzer, Dredze, and
  Pereira}]{blitzer2007biographies}
John Blitzer, Mark Dredze, and Fernando Pereira. 2007.
\newblock Biographies, bollywood, boom-boxes and blenders: Domain adaptation
  for sentiment classification.
\newblock In \emph{Proceedings of the 45th annual meeting of the association of
  computational linguistics}, pages 440--447.

\bibitem[{Boureau et~al.(2010)Boureau, Ponce, and
  LeCun}]{boureau2010theoretical}
Y-Lan Boureau, Jean Ponce, and Yann LeCun. 2010.
\newblock A theoretical analysis of feature pooling in visual recognition.
\newblock In \emph{Proceedings of the 27th international conference on machine
  learning (ICML-10)}, pages 111--118.

\bibitem[{Chollet et~al.(2015)}]{chollet2015keras}
Fran{\c{c}}ois Chollet et~al. 2015.
\newblock Keras.
\newblock \url{https://github.com/keras-team/keras}.

\bibitem[{Coppersmith et~al.(2014)Coppersmith, Dredze, and
  Harman}]{coppersmith2014quantifying}
Glen Coppersmith, Mark Dredze, and Craig Harman. 2014.
\newblock Quantifying mental health signals in twitter.
\newblock In \emph{Proceedings of the Workshop on Computational Linguistics and
  Clinical Psychology: From Linguistic Signal to Clinical Reality}, pages
  51--60.

\bibitem[{De~Choudhury et~al.(2013)De~Choudhury, Counts, and
  Horvitz}]{de2013predicting}
Munmun De~Choudhury, Scott Counts, and Eric Horvitz. 2013.
\newblock Predicting postpartum changes in emotion and behavior via social
  media.
\newblock In \emph{Proceedings of the SIGCHI Conference on Human Factors in
  Computing Systems}, pages 3267--3276. ACM.

\bibitem[{Deriu et~al.(2017)Deriu, Lucchi, De~Luca, Severyn, M{\"u}ller,
  Cieliebak, Hofmann, and Jaggi}]{deriu2017leveraging}
Jan Deriu, Aurelien Lucchi, Valeria De~Luca, Aliaksei Severyn, Simon
  M{\"u}ller, Mark Cieliebak, Thomas Hofmann, and Martin Jaggi. 2017.
\newblock Leveraging large amounts of weakly supervised data for multi-language
  sentiment classification.
\newblock In \emph{Proceedings of the 26th International Conference on World
  Wide Web}, pages 1045--1052. International World Wide Web Conferences
  Steering Committee.

\bibitem[{Eisner et~al.(2016)Eisner, Rockt{\"a}schel, Augenstein,
  Bo{\v{s}}njak, and Riedel}]{eisner2016emoji2vec}
Ben Eisner, Tim Rockt{\"a}schel, Isabelle Augenstein, Matko Bo{\v{s}}njak, and
  Sebastian Riedel. 2016.
\newblock emoji2vec: Learning emoji representations from their description.
\newblock \emph{arXiv preprint arXiv:1609.08359}.

\bibitem[{Ekman(1992)}]{ekman1992argument}
Paul Ekman. 1992.
\newblock An argument for basic emotions.
\newblock \emph{Cognition \& emotion}, 6(3-4):169--200.

\bibitem[{Felbo et~al.(2017)Felbo, Mislove, S{\o}gaard, Rahwan, and
  Lehmann}]{felbo2017using}
Bjarke Felbo, Alan Mislove, Anders S{\o}gaard, Iyad Rahwan, and Sune Lehmann.
  2017.
\newblock Using millions of emoji occurrences to learn any-domain
  representations for detecting sentiment, emotion and sarcasm.
\newblock \emph{arXiv preprint arXiv:1708.00524}.

\bibitem[{Go et~al.(2009)Go, Bhayani, and Huang}]{go2009twitter}
Alec Go, Richa Bhayani, and Lei Huang. 2009.
\newblock Twitter sentiment classification using distant supervision.
\newblock \emph{CS224N Project Report, Stanford}, 1(2009):12.

\bibitem[{Gonz{\'a}lez-Ib{\'a}nez et~al.(2011)Gonz{\'a}lez-Ib{\'a}nez, Muresan,
  and Wacholder}]{gonzalez2011identifying}
Roberto Gonz{\'a}lez-Ib{\'a}nez, Smaranda Muresan, and Nina Wacholder. 2011.
\newblock Identifying sarcasm in twitter: a closer look.
\newblock In \emph{Proceedings of the 49th Annual Meeting of the Association
  for Computational Linguistics: Human Language Technologies: Short
  Papers-Volume 2}, pages 581--586. Association for Computational Linguistics.

\bibitem[{Harman and Dredze(2014)}]{harman2014measuring}
GACCT Harman and Mark~H Dredze. 2014.
\newblock Measuring post traumatic stress disorder in twitter.
\newblock \emph{In ICWSM}.

\bibitem[{Hinton et~al.(2012)Hinton, Srivastava, Krizhevsky, Sutskever, and
  Salakhutdinov}]{hinton2012improving}
Geoffrey~E Hinton, Nitish Srivastava, Alex Krizhevsky, Ilya Sutskever, and
  Ruslan~R Salakhutdinov. 2012.
\newblock Improving neural networks by preventing co-adaptation of feature
  detectors.
\newblock \emph{arXiv preprint arXiv:1207.0580}.

\bibitem[{Ivanov(2017)}]{ilya2017}
Ilya Ivanov. 2017.
\newblock Sentiment bi-rnn.
\newblock \url{https://github.com/ilivans/attention-sentiment}.

\bibitem[{Kim(2014)}]{kim2014convolutional}
Yoon Kim. 2014.
\newblock Convolutional neural networks for sentence classification.
\newblock \emph{arXiv preprint arXiv:1408.5882}.

\bibitem[{Kingma and Ba(2014)}]{kingma2014adam}
Diederik~P Kingma and Jimmy Ba. 2014.
\newblock Adam: A method for stochastic optimization.
\newblock \emph{arXiv preprint arXiv:1412.6980}.

\bibitem[{Liew and Turtle(2016)}]{liew2016exploring}
Jasy Suet~Yan Liew and Howard~R Turtle. 2016.
\newblock Exploring fine-grained emotion detection in tweets.
\newblock In \emph{Proceedings of the NAACL Student Research Workshop}, pages
  73--80.

\bibitem[{Mikolov et~al.(2013)Mikolov, Chen, Corrado, and
  Dean}]{mikolov2013efficient}
Tomas Mikolov, Kai Chen, Greg Corrado, and Jeffrey Dean. 2013.
\newblock Efficient estimation of word representations in vector space.
\newblock \emph{arXiv preprint arXiv:1301.3781}.

\bibitem[{Mintz et~al.(2009)Mintz, Bills, Snow, and
  Jurafsky}]{mintz2009distant}
Mike Mintz, Steven Bills, Rion Snow, and Dan Jurafsky. 2009.
\newblock Distant supervision for relation extraction without labeled data.
\newblock In \emph{Proceedings of the Joint Conference of the 47th Annual
  Meeting of the ACL and the 4th International Joint Conference on Natural
  Language Processing of the AFNLP: Volume 2-Volume 2}, pages 1003--1011.
  Association for Computational Linguistics.

\bibitem[{Mohammad(2012)}]{mohammad2012emotional}
Saif~M Mohammad. 2012.
\newblock Emotional tweets.
\newblock In \emph{Proceedings of the First Joint Conference on Lexical and
  Computational Semantics-Volume 1: Proceedings of the main conference and the
  shared task, and Volume 2: Proceedings of the Sixth International Workshop on
  Semantic Evaluation}, pages 246--255. Association for Computational
  Linguistics.

\bibitem[{Mohammad and Bravo-Marquez(2017)}]{mohammad2017emotion}
Saif~M Mohammad and Felipe Bravo-Marquez. 2017.
\newblock Emotion intensities in tweets.
\newblock \emph{arXiv preprint arXiv:1708.03696}.

\bibitem[{Mohammad and Kiritchenko(2015)}]{mohammad2015using}
Saif~M Mohammad and Svetlana Kiritchenko. 2015.
\newblock Using hashtags to capture fine emotion categories from tweets.
\newblock \emph{Computational Intelligence}, 31(2):301--326.

\bibitem[{Mohammad and Turney(2013)}]{mohammad2013crowdsourcing}
Saif~M Mohammad and Peter~D Turney. 2013.
\newblock Crowdsourcing a word--emotion association lexicon.
\newblock \emph{Computational Intelligence}, 29(3):436--465.

\bibitem[{Nair and Hinton(2010)}]{nair2010rectified}
Vinod Nair and Geoffrey~E Hinton. 2010.
\newblock Rectified linear units improve restricted boltzmann machines.
\newblock In \emph{Proceedings of the 27th international conference on machine
  learning (ICML-10)}, pages 807--814.

\bibitem[{Nguyen and Nguyen(2017)}]{nguyen2017sentence}
Huy-Thanh Nguyen and Minh-Le Nguyen. 2017.
\newblock Sentence modeling with deep neural architecture using lexicon and
  character attention mechanism for sentiment classification.
\newblock In \emph{Proceedings of the Eighth International Joint Conference on
  Natural Language Processing (Volume 1: Long Papers)}, volume~1, pages
  536--544.

\bibitem[{Pang et~al.(2002)Pang, Lee, and Vaithyanathan}]{pang2002thumbs}
Bo~Pang, Lillian Lee, and Shivakumar Vaithyanathan. 2002.
\newblock Thumbs up?: sentiment classification using machine learning
  techniques.
\newblock In \emph{Proceedings of the ACL-02 conference on Empirical methods in
  natural language processing-Volume 10}, pages 79--86. Association for
  Computational Linguistics.

\bibitem[{Park et~al.(2012)Park, Cha, and Cha}]{park2012depressive}
Minsu Park, Chiyoung Cha, and Meeyoung Cha. 2012.
\newblock Depressive moods of users portrayed in twitter.
\newblock In \emph{Proceedings of the ACM SIGKDD Workshop on healthcare
  informatics (HI-KDD)}, volume 2012, pages 1--8. ACM New York, NY.

\bibitem[{Pennebaker et~al.(2007)Pennebaker, Booth, and
  Francis}]{pennebaker2007linguistic}
James~W Pennebaker, Roger~J Booth, and Martha~E Francis. 2007.
\newblock Linguistic inquiry and word count: Liwc [computer software].
\newblock \emph{Austin, TX: liwc. net}.

\bibitem[{Plutchik(2001)}]{plutchik2001nature}
Robert Plutchik. 2001.
\newblock The nature of emotions human emotions have deep evolutionary roots, a
  fact that may explain their complexity and provide tools for clinical
  practice.
\newblock \emph{American scientist}, 89(4):344--350.

\bibitem[{Poria et~al.(2016)Poria, Chaturvedi, Cambria, and
  Hussain}]{poria2016convolutional}
Soujanya Poria, Iti Chaturvedi, Erik Cambria, and Amir Hussain. 2016.
\newblock Convolutional mkl based multimodal emotion recognition and sentiment
  analysis.
\newblock In \emph{Data Mining (ICDM), 2016 IEEE 16th International Conference
  on}, pages 439--448. IEEE.

\bibitem[{Purver and Battersby(2012)}]{purver2012experimenting}
Matthew Purver and Stuart Battersby. 2012.
\newblock Experimenting with distant supervision for emotion classification.
\newblock In \emph{Proceedings of the 13th Conference of the European Chapter
  of the Association for Computational Linguistics}, pages 482--491.
  Association for Computational Linguistics.

\bibitem[{Qadir and Riloff(2013)}]{qadir2013bootstrapped}
Ashequl Qadir and Ellen Riloff. 2013.
\newblock Bootstrapped learning of emotion hashtags\# hashtags4you.
\newblock In \emph{Proceedings of the 4th workshop on computational approaches
  to subjectivity, sentiment and social media analysis}, pages 2--11.

\bibitem[{Read(2005)}]{read2005using}
Jonathon Read. 2005.
\newblock Using emoticons to reduce dependency in machine learning techniques
  for sentiment classification.
\newblock In \emph{Proceedings of the ACL student research workshop}, pages
  43--48. Association for Computational Linguistics.

\bibitem[{Riloff(1996)}]{riloff1996automatically}
Ellen Riloff. 1996.
\newblock Automatically generating extraction patterns from untagged text.
\newblock In \emph{Proceedings of the national conference on artificial
  intelligence}, pages 1044--1049.

\bibitem[{Riloff and Wiebe(2003)}]{riloff2003learning}
Ellen Riloff and Janyce Wiebe. 2003.
\newblock Learning extraction patterns for subjective expressions.
\newblock In \emph{Proceedings of the 2003 conference on Empirical methods in
  natural language processing}, pages 105--112. Association for Computational
  Linguistics.

\bibitem[{Roberts et~al.(2012)Roberts, Roach, Johnson, Guthrie, and
  Harabagiu}]{roberts2012empatweet}
Kirk Roberts, Michael~A Roach, Joseph Johnson, Josh Guthrie, and Sanda~M
  Harabagiu. 2012.
\newblock Empatweet: Annotating and detecting emotions on twitter.
\newblock In \emph{LREC}, volume~12, pages 3806--3813.

\bibitem[{Rosenthal et~al.(2017)Rosenthal, Farra, and
  Nakov}]{SemEval:2017:task4}
Sara Rosenthal, Noura Farra, and Preslav Nakov. 2017.
\newblock {SemEval}-2017 task 4: Sentiment analysis in {T}witter.
\newblock In \emph{Proceedings of the 11th International Workshop on Semantic
  Evaluation}, SemEval '17, Vancouver, Canada. Association for Computational
  Linguistics.

\bibitem[{Santos et~al.(2017)Santos, Corr{\^e}a~Jr, Oliveira~Jr, Amancio,
  Mansur, and Alu{\'\i}sio}]{santos2017enriching}
Leandro B~dos Santos, Edilson~A Corr{\^e}a~Jr, Osvaldo~N Oliveira~Jr, Diego~R
  Amancio, Let{\'\i}cia~L Mansur, and Sandra~M Alu{\'\i}sio. 2017.
\newblock Enriching complex networks with word embeddings for detecting mild
  cognitive impairment from speech transcripts.
\newblock \emph{arXiv preprint arXiv:1704.08088}.

\bibitem[{Sap et~al.(2014)Sap, Park, Eichstaedt, Kern, Stillwell, Kosinski,
  Ungar, and Schwartz}]{sap2014developing}
Maarten Sap, Gregory Park, Johannes Eichstaedt, Margaret Kern, David Stillwell,
  Michal Kosinski, Lyle Ungar, and Hansen~Andrew Schwartz. 2014.
\newblock Developing age and gender predictive lexica over social media.
\newblock In \emph{Proceedings of the 2014 Conference on Empirical Methods in
  Natural Language Processing (EMNLP)}, pages 1146--1151.

\bibitem[{Savigny and Purwarianti(2017)}]{savigny2017emotion}
Julio Savigny and Ayu Purwarianti. 2017.
\newblock Emotion classification on youtube comments using word embedding.
\newblock In \emph{Advanced Informatics, Concepts, Theory, and Applications
  (ICAICTA), 2017 International Conference on}, pages 1--5. IEEE.

\bibitem[{Sintsova et~al.(2013)Sintsova, Musat, and Pu}]{sintsova2013fine}
Valentina Sintsova, Claudiu-Cristian Musat, and Pearl Pu. 2013.
\newblock Fine-grained emotion recognition in olympic tweets based on human
  computation.
\newblock In \emph{4th Workshop on computational approaches to subjectivity,
  sentiment and social media analysis}, EPFL-CONF-197185.

\bibitem[{Socher et~al.(2013)Socher, Perelygin, Wu, Chuang, Manning, Ng, and
  Potts}]{socher2013recursive}
Richard Socher, Alex Perelygin, Jean Wu, Jason Chuang, Christopher~D Manning,
  Andrew Ng, and Christopher Potts. 2013.
\newblock Recursive deep models for semantic compositionality over a sentiment
  treebank.
\newblock In \emph{Proceedings of the 2013 conference on empirical methods in
  natural language processing}, pages 1631--1642.

\bibitem[{Strapparava and Mihalcea(2007)}]{strapparava2007semeval}
Carlo Strapparava and Rada Mihalcea. 2007.
\newblock Semeval-2007 task 14: Affective text.
\newblock In \emph{Proceedings of the 4th International Workshop on Semantic
  Evaluations}, pages 70--74. Association for Computational Linguistics.

\bibitem[{Strapparava et~al.(2004)Strapparava, Valitutti
  et~al.}]{strapparava2004wordnet}
Carlo Strapparava, Alessandro Valitutti, et~al. 2004.
\newblock Wordnet affect: an affective extension of wordnet.
\newblock In \emph{LREC}, volume~4, pages 1083--1086.

\bibitem[{Suttles and Ide(2013)}]{suttles2013distant}
Jared Suttles and Nancy Ide. 2013.
\newblock Distant supervision for emotion classification with discrete binary
  values.
\newblock In \emph{International Conference on Intelligent Text Processing and
  Computational Linguistics}, pages 121--136. Springer.

\bibitem[{Tromp and Pechenizkiy(2015)}]{tromp2015pattern}
Erik Tromp and Mykola Pechenizkiy. 2015.
\newblock Pattern-based emotion classification on social media.
\newblock In \emph{Advances in Social Media Analysis}, pages 1--20. Springer.

\bibitem[{Volkova and Bachrach(2016)}]{volkova2016inferring}
Svitlana Volkova and Yoram Bachrach. 2016.
\newblock Inferring perceived demographics from user emotional tone and
  user-environment emotional contrast.
\newblock In \emph{Proceedings of the 54th Annual Meeting of the Association
  for Computational Linguistics (Volume 1: Long Papers)}, volume~1, pages
  1567--1578.

\bibitem[{Volkova et~al.(2013)Volkova, Wilson, and
  Yarowsky}]{volkova2013exploring}
Svitlana Volkova, Theresa Wilson, and David Yarowsky. 2013.
\newblock Exploring sentiment in social media: Bootstrapping subjectivity clues
  from multilingual twitter streams.
\newblock In \emph{Proceedings of the 51st Annual Meeting of the Association
  for Computational Linguistics (Volume 2: Short Papers)}, volume~2, pages
  505--510.

\bibitem[{Wallbott and Scherer(1988)}]{wallbott1988universal}
Harald~G Wallbott and Klaus~R Scherer. 1988.
\newblock How universal and specific is emotional experience?: Evidence from 27
  countries on five continents.

\bibitem[{Wang et~al.(2012)Wang, Chen, Thirunarayan, and
  Sheth}]{wang2012harnessing}
Wenbo Wang, Lu~Chen, Krishnaprasad Thirunarayan, and Amit~P Sheth. 2012.
\newblock Harnessing twitter" big data" for automatic emotion identification.
\newblock In \emph{Privacy, Security, Risk and Trust (PASSAT), 2012
  International Conference on and 2012 International Confernece on Social
  Computing (SocialCom)}, pages 587--592. IEEE.

\bibitem[{Ward~Jr(1963)}]{ward1963hierarchical}
Joe~H Ward~Jr. 1963.
\newblock Hierarchical grouping to optimize an objective function.
\newblock \emph{Journal of the American statistical association},
  58(301):236--244.

\bibitem[{Weidman et~al.(2017)Weidman, Steckler, and Tracy}]{weidman2017jingle}
Aaron~C Weidman, Conor~M Steckler, and Jessica~L Tracy. 2017.
\newblock The jingle and jangle of emotion assessment: Imprecise measurement,
  casual scale usage, and conceptual fuzziness in emotion research.
\newblock \emph{Emotion}, 17(2):267.

\bibitem[{Zhang et~al.(2015)Zhang, Zhao, and LeCun}]{zhang2015character}
Xiang Zhang, Junbo Zhao, and Yann LeCun. 2015.
\newblock Character-level convolutional networks for text classification.
\newblock In \emph{Advances in neural information processing systems}, pages
  649--657.

\bibitem[{Zhou et~al.(2017)Zhou, Huang, Zhang, Zhu, and
  Liu}]{zhou2017emotional}
Hao Zhou, Minlie Huang, Tianyang Zhang, Xiaoyan Zhu, and Bing Liu. 2017.
\newblock Emotional chatting machine: emotional conversation generation with
  internal and external memory.
\newblock \emph{arXiv preprint arXiv:1704.01074}.

\bibitem[{ZOLKEPLI(2017)}]{husein}
HUSEIN ZOLKEPLI. 2017.
\newblock Emotion classification.
\newblock
  \url{https://github.com/huseinzol05/Emotion-Classification-Comparison}.

\end{thebibliography}
\bibliographystyle{acl_natbib}

\end{document}